\PassOptionsToPackage{unicode}{hyperref}
\PassOptionsToPackage{hyphens}{url}
\documentclass[
]{article}
\usepackage{xcolor}
\usepackage[margin=2cm]{geometry}
\usepackage{amsmath,amssymb}
\usepackage{iftex}
\ifPDFTeX
  \usepackage[T1]{fontenc}
  \usepackage[utf8]{inputenc}
  \usepackage{textcomp} 
\else 
  \usepackage{unicode-math} 
  \defaultfontfeatures{Scale=MatchLowercase}
  \defaultfontfeatures[\rmfamily]{Ligatures=TeX,Scale=1}
\fi
\usepackage{lmodern}
\ifPDFTeX\else
\fi
\IfFileExists{upquote.sty}{\usepackage{upquote}}{}
\IfFileExists{microtype.sty}{
  \usepackage[]{microtype}
  \UseMicrotypeSet[protrusion]{basicmath} 
}{}
\makeatletter
\@ifundefined{KOMAClassName}{
  \IfFileExists{parskip.sty}{%
    \usepackage{parskip}
  }{
    \setlength{\parindent}{0pt}
    \setlength{\parskip}{6pt plus 2pt minus 1pt}}
}{
  \KOMAoptions{parskip=half}}
\makeatother
\usepackage{longtable,booktabs,array}
\usepackage{multirow}
\usepackage{calc} 
\usepackage{etoolbox}
\makeatletter
\patchcmd\longtable{\par}{\if@noskipsec\mbox{}\fi\par}{}{}
\makeatother
\IfFileExists{footnotehyper.sty}{\usepackage{footnotehyper}}{\usepackage{footnote}}
\makesavenoteenv{longtable}
\usepackage{graphicx}
\makeatletter
\newsavebox\pandoc@box
\newcommand*\pandocbounded[1]{
  \sbox\pandoc@box{#1}%
  \Gscale@div\@tempa{\textheight}{\dimexpr\ht\pandoc@box+\dp\pandoc@box\relax}%
  \Gscale@div\@tempb{\linewidth}{\wd\pandoc@box}%
  \ifdim\@tempb\p@<\@tempa\p@\let\@tempa\@tempb\fi
  \ifdim\@tempa\p@<\p@\scalebox{\@tempa}{\usebox\pandoc@box}%
  \else\usebox{\pandoc@box}%
  \fi%
}
\def\fps@figure{htbp}
\makeatother
\NewDocumentCommand\citeproctext{}{}

\makeatletter
 \let\@cite@ofmt\@firstofone
 \def\@biblabel#1{}
 \def\@cite#1#2{{#1\if@tempswa , #2\fi}}
\makeatother
\newlength{\cslhangindent}
\newlength{\csllabelwidth}
\newenvironment{CSLReferences}[2] 
 {\begin{list}{}{%
  \setlength{\itemindent}{0pt}
  \setlength{\leftmargin}{0pt}
  \setlength{\parsep}{0pt}
  \ifodd #1
   \setlength{\leftmargin}{\cslhangindent}
   \setlength{\itemindent}{-1\cslhangindent}
  \fi
  \setlength{\itemsep}{#2\baselineskip}}}
 {\end{list}}
\usepackage{calc}

\newcommand{\CSLLeftMargin}[1]{\parbox[t]{\csllabelwidth}{\strut#1\strut}}
\newcommand{\CSLRightInline}[1]{\parbox[t]{\linewidth - \csllabelwidth}{\strut#1\strut}}

\usepackage{etoolbox}
\makeatletter\patchcmd{\@maketitle}{\LARGE}{\LARGE\bfseries}{}{}\makeatother
\usepackage{caption}
\AtBeginEnvironment{abstract}{}
\usepackage{authblk}
\renewcommand\Authfont{\large}
\renewcommand\Affilfont{\small}
\renewcommand\Authands{, }
\renewcommand\Authand{, }
\renewcommand{\thefootnote}{\fnsymbol{footnote}}
\author[1,2]{J. Raphael Schäfer}
\author[1]{Kai Geissler}
\author[1]{Till Nicke}
\author[1]{Chiara Tappermann}
\author[1]{Karoline Heber}
\author[1,3]{Eike Petersen}
\author[1]{Habib Mergan}
\author[1]{Lars Ole Schwen}
\author[1]{Nick Weiss}
\author[1]{Annika Gerken}
\author[1]{Jan Hendrik Moltz}
\author[4,5]{Tom Bisson}
\author[5]{Isil Dogan O}
\author[5]{Tim-Rasmus Kiehl}
\author[5]{Norman Zerbe}
\author[6]{Sefer Elezkurtaj}
\author[7]{Robin S. Mayer}
\author[7]{Nadine Flinner}
\author[7]{Peter Wild}
\author[8]{Isabel Dahm}
\author[8]{Felix Peisen}
\author[9]{Heinrich von Busch}
\author[10]{Robert Grimm}
\author[11,12]{Sebastian Arndt}
\author[11,13]{Lisa Siegler}
\author[11,13]{Matthias Stefan May}
\author[14]{Antje Prasse}
\author[14]{Natalia Artysh}
\author[1,2]{Fabian Kiessling\thanks{equal contribution}}
\author[1]{Johannes Lotz\textsuperscript{\textdagger}}
\affil[1]{Fraunhofer Institute for Digital Medicine MEVIS}
\affil[2]{Institute for Experimental Molecular Imaging, RWTH Aachen University, Aachen, Germany}
\affil[3]{Institute of Diagnostic and Interventional Radiology, Medizinische Hochschule Hannover, Hannover, Germany}
\affil[4]{Department of Pathology, Massachusetts General Hospital and Harvard Medical School, Boston, MA, USA}
\affil[5]{Institute of Medical Informatics, Charité – Universitätsmedizin Berlin, corporate member of Freie Universität Berlin und Humboldt-Universität zu Berlin, Berlin, Germany}
\affil[6]{Institute of Pathology, Charité – Universitätsmedizin Berlin, corporate member of Freie Universität Berlin und Humboldt-Universität zu Berlin, Berlin, Germany}
\affil[7]{Dr. Senckenberg Institutes of Pathology, Neuropathology and Human Genetics, Goethe University Frankfurt and University Medical Center Frankfurt, Frankfurt am Main, Germany}
\affil[8]{Department of Diagnostic and Interventional Radiology, Tübingen University Hospital, Tübingen, Germany}
\affil[9]{Digital \& Automation, Siemens Healthineers AG, Forchheim, Germany}
\affil[10]{Research \& Clinical Translation for Magnetic Resonance, Siemens Healthineers AG, Erlangen, Germany}
\affil[11]{Department of Radiology, University Hospital Erlangen and Friedrich-Alexander-Universität Erlangen-Nürnberg, Erlangen, Germany}
\affil[12]{Medical Centre for Information and Communication Technology, University Hospital Erlangen, Erlangen, Germany}
\affil[13]{Imaging Science Institute, Uniklinik Erlangen, Erlangen, Germany}
\affil[14]{Fraunhofer Institute for Toxicology and Experimental Medicine ITEM, Hannover, Germany}
\renewcommand{\author}[2][]{}
\makeatletter
\@ifpackageloaded{subcaption}{}{\usepackage{subcaption}}
\@ifpackageloaded{caption}{}{\usepackage{caption}}
\AtBeginDocument{%

}
\AtBeginDocument{%

}
\newcounter{pandoccrossref@subfigures@footnote@counter}
{\end{figure}%
\addtocounter{footnote}{-\value{pandoccrossref@subfigures@footnote@counter}}
\@for\f:=\global@pandoccrossref@subfigures@footnotes\do{\stepcounter{footnote}\footnotetext{\f}}%
\gdef\global@pandoccrossref@subfigures@footnotes{}}
\@ifpackageloaded{float}{}{\usepackage{float}}
\floatstyle{ruled}
\@ifundefined{c@chapter}{\newfloat{codelisting}{h}{lop}}{\newfloat{codelisting}{h}{lop}[chapter]}
\floatname{codelisting}{Listing}

\makeatother
\usepackage{bookmark}
\IfFileExists{xurl.sty}{\usepackage{xurl}}{} 
\hypersetup{
  pdftitle={CoM³eT: A foundation model for medical image analysis through federated{,} multidimensional context integration},
  hidelinks,
  pdfcreator={LaTeX via pandoc}}

\title{CoM³eT: A foundation model for medical image analysis through
federated, multidimensional context integration}
\author{}
\date{}

\begin{document}
\maketitle
\begin{abstract}
Medical foundation models improve generalization when training AI models
with limited labeled data, but remain confined to a single specialty,
such as pathology or radiology, and to either sparse or dense outputs,
such as classification or segmentation. Here, we present CoM³eT
(\textbf{Co}-representation \textbf{M}ultidimensional \textbf{M}ultitask
\textbf{Me}dical \textbf{T}ransformer), a medical vision foundation
model that unifies pathology and radiology, sparse and dense
predictions, and two- and higher-dimensional inputs by modeling
multidimensional context with attention. CoM³eT outperformed other
medical foundation models in an open competition spanning five
tomographic, four whole-specimen, and three two-dimensional datasets,
covering sparse and dense prediction tasks as well as report generation.
When adapted across diverse clinical applications, training fewer than
2.5\% of parameters achieved performance comparable to full fine-tuning,
enabling research without access to high-performance GPU clusters.
Applied to federated learning across hospitals, this approach achieved
performance comparable to pooled-data training over internet connections
and with consumer-grade hardware.
\end{abstract}

\setcounter{footnote}{0}\renewcommand{\thefootnote}{\arabic{footnote}}

\pagebreak

\subsection{Main}\label{main}

Artificial intelligence (AI) is improving the analysis of medical data,
enabling more accurate diagnoses and more personalized therapy
decisions. In particular, deep learning can turn massive datasets into
multi-purpose models. However, high-quality, large-scale imaging
datasets are unavailable for many diseases, which limits model
generalization.

To address data scarcity, vision foundation models (FMs) can reduce
task-specific data requirements by pretraining on diverse datasets. For
example, VIRCHOW\textsuperscript{1} is a vision transformer trained on
1.5 million histopathology images from a single modality and stain, and
CT-FM\textsuperscript{2} performs strongly across multiple CT tasks
after pretraining on 148,000 CTs. Such medical FMs are typically limited
to single specialties such as pathology or radiology, or even single
modalities, and thus cannot jointly analyze a patient's multimodal
images\textsuperscript{3}. The limited scope of medical FMs could stem
from the difficulty of curating general pretraining datasets. Datasets
from textbooks (e.g., PathCAP\textsuperscript{4}) or social media (e.g.,
PLIP\textsuperscript{5}) may not align with clinical practice due to
selection biases. Alternatively, datasets mined directly from clinical
information systems achieve large quantities and full image quality, but
often originate from one or a few sites\textsuperscript{1,2}. Because
data must be pooled for pretraining, these datasets tend to be narrow
and fail to capture the diversity of global populations.

Federated learning lets institutions train models collaboratively
without pooling data. This can represent population diversity better and
reduce the need for central data-pooling infrastructure. However, large
models make this difficult. For example, federated
averaging\textsuperscript{6} requires repeated exchange of the full
model between participating sites and a central server, which is
impractical for large models. Parameter-efficient fine-tuning addresses
this by reducing the number of trainable parameters while achieving
near-centralized performance\textsuperscript{7}. However, if the
fine-tuned parameters are distributed throughout the model, the full
computation must run at every site, requiring powerful hardware
everywhere. Instead, partial fine-tuning updates only top layers while
keeping the base of the model frozen, enabling most of the computation
to be performed only once.

Besides being restricted in data origin, existing models are typically
applicable to either sparse tasks such as classification and regression,
or dense tasks such as segmentation and instance detection, whereas
general medical vision FMs require both. UNI\textsuperscript{8} is a
vision transformer for computational pathology that extends to
segmentation by adding dense adapters. However, due to the
non-hierarchical nature of transformers, segmentation performance lags
behind specialized architectures, requiring a relatively complex dense
adapter and a large number of segmentation-specific parameters (up to
50\%), which makes them less data-efficient\textsuperscript{9}. In
segmentation, nnU-Net\textsuperscript{10} has demonstrated the continued
strength of the traditional hierarchical U-Net but it uses a manually
defined pipeline with slightly different architectures for different
tasks and is therefore not universally pretrainable.

Beyond being limited to either sparse or dense tasks, handling many data
types challenges medical vision FMs. Medical images come in diverse
formats, including high-resolution 2D images (e.g.~X-rays and tissue
microarrays), gigapixel images (e.g.~whole-slide images in pathology),
and volumetric images (e.g.~CT and MRI scans). A solution is to
decompose them into their greatest common denominator: 2D image patches,
such as frames in videos, slices in tomographic images, and tiles in
gigapixel images. For example, SLIViT\textsuperscript{11} approached 3D
classification using a 2D encoder for feature extraction and a
transformer for aggregation. Similarly, CAT-Net\textsuperscript{12}
introduced a cross-slice transformer for 3D segmentation that, unlike 3D
networks, maintained its performance on flat/anisotropic volumes while
enabling long-range information exchange between slices.

In this article, we show that a universal medical vision FM for
multidimensional images (CoM³eT; \textbf{Co}-representation
\textbf{M}ultidimensional \textbf{M}ultitask \textbf{Me}dical
\textbf{T}ransformer) can be trained by combining a strong image encoder
with transformer modules. While CAT-Net is limited to segmentation and
SLIViT to classification, CoM³eT processes both for multidimensional
images using co-representations of semantic and spatial information,
referred to as patch and pixel tokens.

\begin{figure}
\centering
\includegraphics[width=0.8\linewidth,height=\textheight,keepaspectratio,alt={Study overview. A, CoM³eT's pretraining includes multidimensional images from multiple medical imaging modalities together with natural image data into a single model via multitask learning. B, the image transformer models global context from outputs of the vision backbone. C, the pyramid transformer uses this context to extend FM segmentation to multidimensional images such as 3D volumes. D-F, pretraining uses supervised labels across vision-language, classification and segmentation tasks. G, CoM³eT participated in UNICORN, the first benchmark competition for medical FMs, which compared frozen FMs on a diverse set of medical imaging tasks. H, performance comparison between CoM³eT, a specialized reference model, and CoM³eT restricted to individual patches. I, CoM³eT's architecture supports partial fine-tuning. We compare full and partial fine-tuning and apply partial fine-tuning in a real-world federated learning setting.}]{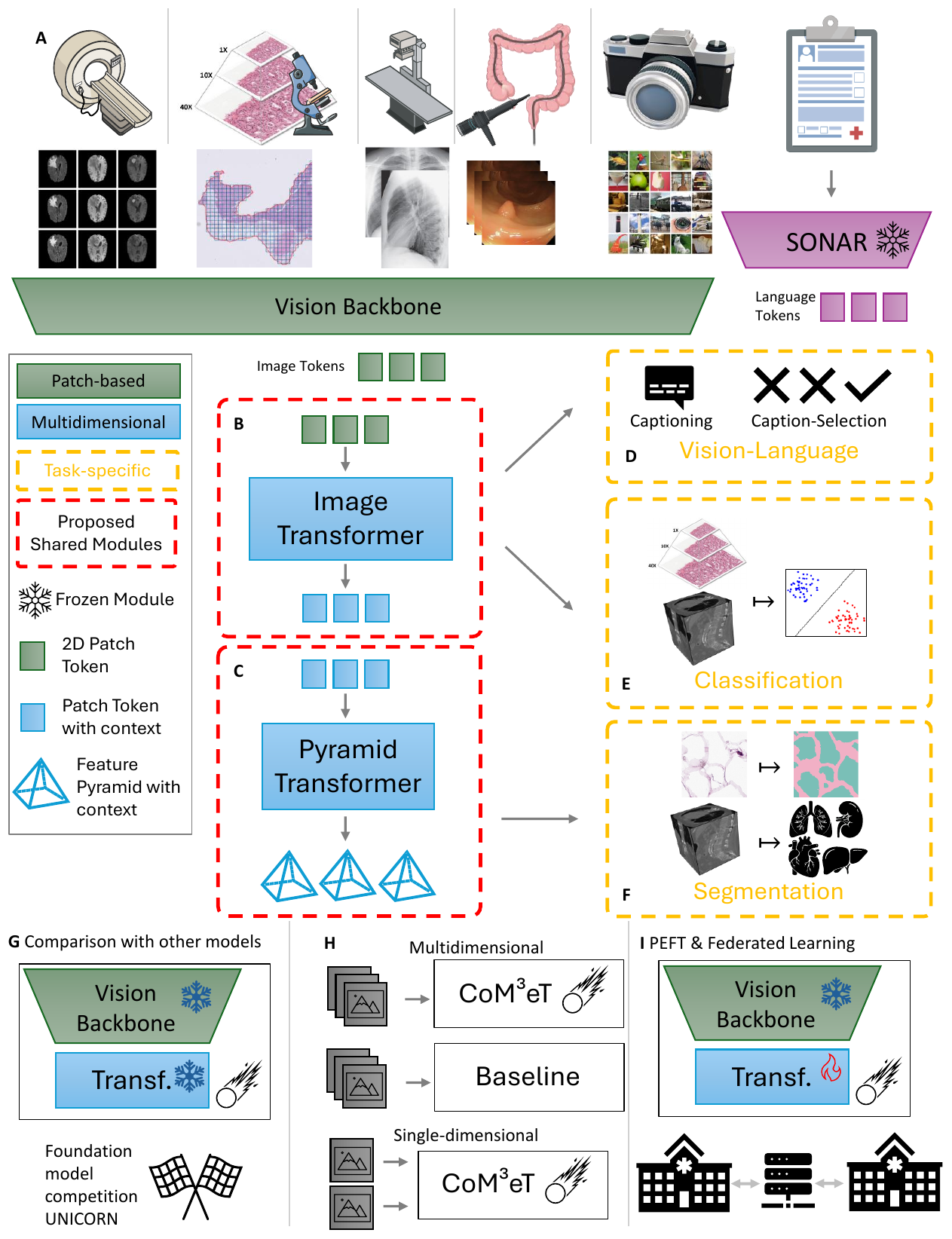}
\caption{\textbf{Study overview.} \textbf{A,} CoM³eT's pretraining
includes multidimensional images from multiple medical imaging
modalities together with natural image data into a single model via
multitask learning. \textbf{B,} the image transformer models global
context from outputs of the vision backbone. \textbf{C,} the pyramid
transformer uses this context to extend FM segmentation to
multidimensional images such as 3D volumes. \textbf{D-F,} pretraining
uses supervised labels across vision-language, classification and
segmentation tasks. \textbf{G,} CoM³eT participated in UNICORN, the
first benchmark competition for medical FMs, which compared frozen FMs
on a diverse set of medical imaging tasks. \textbf{H,} performance
comparison between CoM³eT, a specialized reference model, and CoM³eT
restricted to individual patches. \textbf{I,} CoM³eT's architecture
supports partial fine-tuning. We compare full and partial fine-tuning
and apply partial fine-tuning in a real-world federated learning
setting.}\label{fig:overview}
\end{figure}

First, we constructed CoM³eT by scaling multitask
pretraining\textsuperscript{13} to over 60 tasks. We introduced
attention-based components that handled both sparse tasks (e.g.,
classification) and dense tasks (e.g., segmentation) for
multidimensional images in one model while sharing most parameters
across tasks. Finally, we developed dimensionality-agnostic positional
encoding for multitask FMs. For each contribution, we constructed a
synthetic benchmark to isolate its effect and quantified its impact on
real multidimensional images with a representative test task. The study
is summarized in Fig.~\ref{fig:overview}. To compare with
state-of-the-art FMs, we submitted CoM³eT to the UNICORN competition for
medical FMs, where it ranked first overall and first in both radiology
and pathology\textsuperscript{14}. Then, we showed that partial
fine-tuning was effective even for applications dissimilar to the
pretraining tasks. By reducing training resources and communication
load, partial fine-tuning enabled federated fine-tuning, which we
demonstrated in a real-world setup with German university hospitals.

\subsection{Results}\label{results}

\subsubsection{CoM³eT unifies multidimensional medical imaging tasks in
a single model}\label{sec:results.architecture}

CoM³eT reuses shared modules across tasks. The largest is the vision
backbone, shared by all tasks. It processes image inputs and produces
patch tokens, which can represent full 2D medical images such as X-rays,
frames in videos, slices in tomographic images, or patches in gigapixel
images. Patch tokens are CoM³eT's fundamental units, and groups of patch
tokens represent multidimensional inputs. For segmentation tasks,
semantic information in the patch tokens alone is not sufficient because
segmentation also requires spatial information. Therefore, the vision
backbone also produces feature pyramids, i.e., multi-scale feature maps
that preserve spatial layout across resolutions, as is common in
segmentation architectures.

\begin{figure}
\centering
\includegraphics[width=0.8\linewidth,height=\textheight,keepaspectratio,alt={Model overview. A, First, CoM³eT's vision backbone encodes image inputs, representing a multidimensional medical image as a group of patch tokens and feature pyramids. In the case of multidimensional images, the image transformer then computes global context. For multidimensional segmentation tasks, the pyramid transformer uses the image transformer's output tokens to enrich the feature pyramids with global context. B, CoM³eT creates two types of features. Segmentation tasks use features with their spatial shape intact (hyperpixel tokens), while classification tasks use feature vectors (patch tokens). C, The image transformer is an encoder-only transformer that models interactions between patch tokens to add global multidimensional context while preserving one representation per input. D, The pyramid transformer uses the outputs of the image transformer to inject global multidimensional context into feature maps.}]{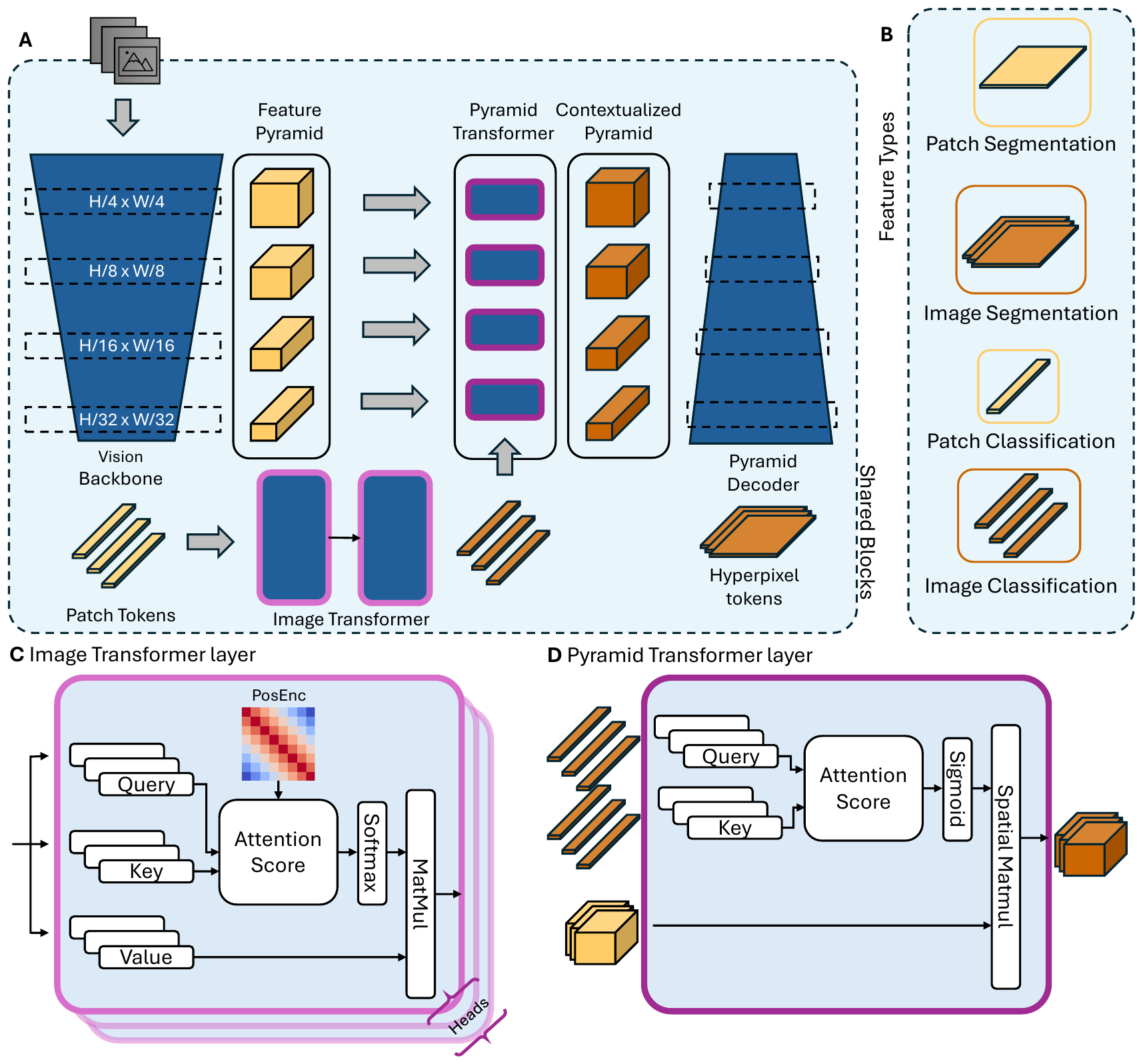}
\caption{\textbf{Model overview.} \textbf{A,} First, CoM³eT's vision
backbone encodes image inputs, representing a multidimensional medical
image as a group of patch tokens and feature pyramids. In the case of
multidimensional images, the image transformer then computes global
context. For multidimensional segmentation tasks, the pyramid
transformer uses the image transformer's output tokens to enrich the
feature pyramids with global context. \textbf{B,} CoM³eT creates two
types of features. Segmentation tasks use features with their spatial
shape intact (hyperpixel tokens), while classification tasks use feature
vectors (patch tokens). \textbf{C,} The image transformer is an
encoder-only transformer that models interactions between patch tokens
to add global multidimensional context while preserving one
representation per input. \textbf{D,} The pyramid transformer uses the
outputs of the image transformer to inject global multidimensional
context into feature maps.}\label{fig:architecture}
\end{figure}

The image transformer models interactions among sets of patch tokens
while preserving one contextualized patch token per input image patch.
This lets the model solve both patch-level tasks, such as identifying
slices that contain tumor, and patient-level tasks, such as cancer
grading. The pyramid transformer extends the image transformer to dense
tasks and uses its output tokens to enrich the vision backbone's feature
maps with global context. Subsequently, the decoder converts these into
fine-grained hyperpixel tokens. Thanks to this modularity, CoM³eT can
support both patch-based and multidimensional imaging data, as well as
sparse and dense prediction tasks. Table~\ref{tbl:architecture} lists
sizes of the model variants, Fig.~\ref{fig:architecture} provides an
overview of the architecture, and
\hyperref[extfig:syndata]{Extended~Data~Fig.~1} illustrates how CoM³eT
was verified to use local and global context and positional information
for classification and segmentation on synthetic data.

\subsubsection{CoM³eT leads the UNICORN
competition}\label{sec:results.unicorn}

\begin{figure}
\centering
\pandocbounded{\includegraphics[keepaspectratio,alt={Results overview. A, Mean task-specific metrics from the UNICORN competition (each task's metric scaled to the range from random-guessing performance to maximum performance). Left, CoM³eT compared with a theoretical best ensemble composed of the strongest competing FM for each task. Middle and right, pathology and radiology results, respectively, compared with the second- and third-place FMs or combinations of FMs. B, Fine-tuning results with partial and full fine-tuning. C, GPU time, measured on an A100, for different fine-tuning settings. For multidimensional images with between 1 and 32 patch tokens, training based on cached patch tokens (backbone caching) has almost zero cost for partial fine-tuning (black). The blue and orange lines overlap because partial fine-tuning and no fine-tuning incur nearly identical cost. D, GPU time for multidimensional images with 64 or more patch tokens. Full fine-tuning cannot be used for these images. E, Time required per update for each fine-tuning setting for CoM³eT and its larger variant, CoM³eT-Large. In federated learning, a full update step requires exchanging the full model state twice, first from the sites to the server and then from the server back to the sites. Unless stated otherwise, ``CoM³eT'' refers to the CoM³eT-Base variant.}]{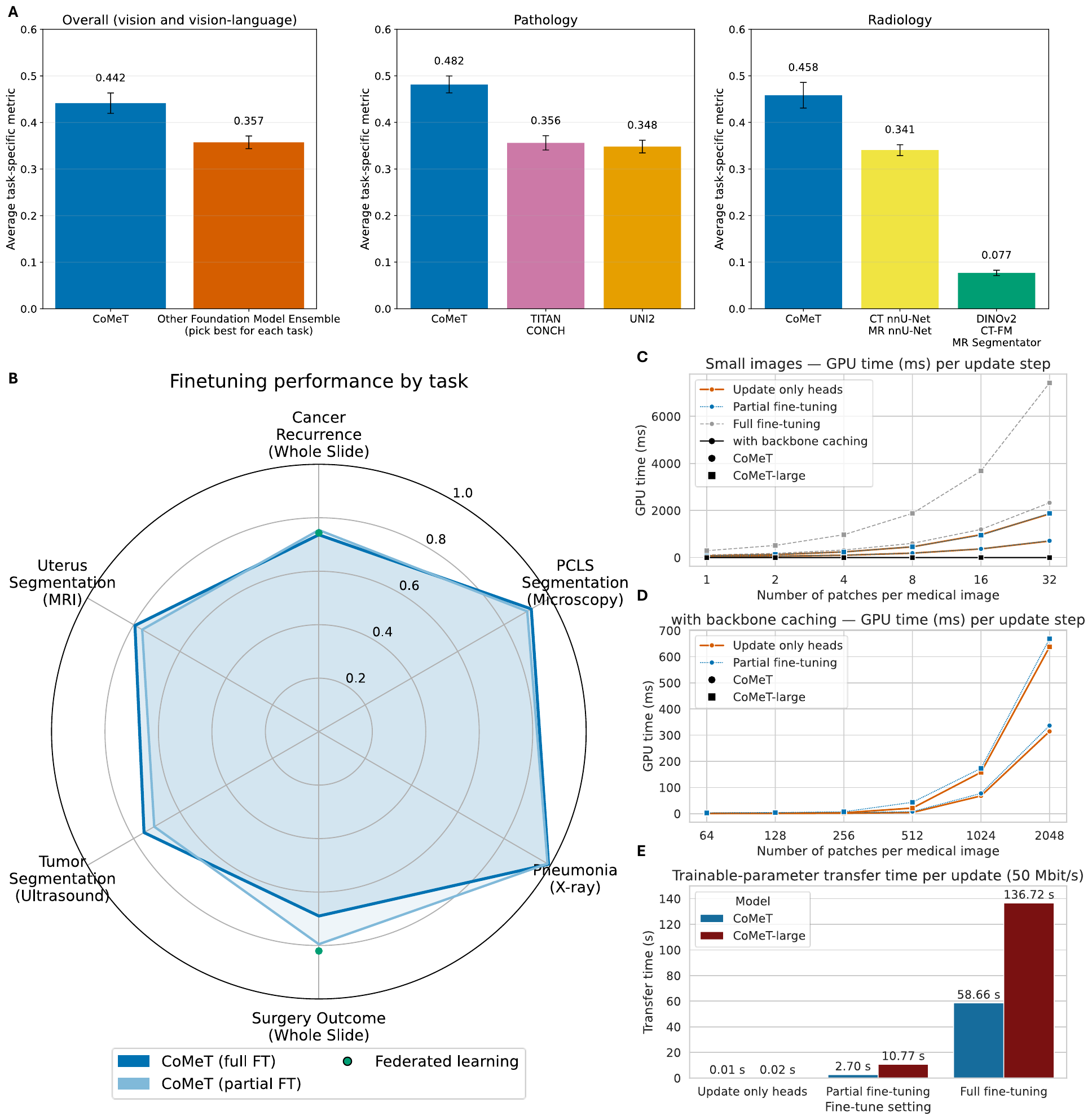}}
\caption{\textbf{Results overview.} \textbf{A,} Mean task-specific
metrics from the UNICORN competition (each task's metric scaled to the
range from random-guessing performance to maximum performance). Left,
CoM³eT compared with a theoretical best ensemble composed of the
strongest competing FM for each task. Middle and right, pathology and
radiology results, respectively, compared with the second- and
third-place FMs or combinations of FMs. \textbf{B,} Fine-tuning results
with partial and full fine-tuning. \textbf{C,} GPU time, measured on an
A100, for different fine-tuning settings. For multidimensional images
with between 1 and 32 patch tokens, training based on cached patch
tokens (backbone caching) has almost zero cost for partial fine-tuning
(black). The blue and orange lines overlap because partial fine-tuning
and no fine-tuning incur nearly identical cost. \textbf{D,} GPU time for
multidimensional images with 64 or more patch tokens. Full fine-tuning
cannot be used for these images. \textbf{E,} Time required per update
for each fine-tuning setting for CoM³eT and its larger variant,
CoM³eT-Large. In federated learning, a full update step requires
exchanging the full model state twice, first from the sites to the
server and then from the server back to the sites. Unless stated
otherwise, ``CoM³eT'' refers to the CoM³eT-Base
variant.}\label{fig:pretrain-results}
\end{figure}

Meaningful comparisons between medical FMs require an independent
benchmark based entirely on private data\textsuperscript{15}, evaluated
without internet access to avoid data leakage. Such a benchmark should
cover the main medical imaging task types, including classification,
pixel-level prediction such as segmentation, and vision-language
reporting, as well as the main imaging domains: pathology with gigapixel
images or large 2D tiles, and radiology with 3D images such as CT and
multiparametric MRI. UNICORN\textsuperscript{14} is the first benchmark
fulfilling these criteria. It uses a frozen setting: model weights
remain fixed and only lightweight adapters are trained on extracted
representations.

Across the benchmark, CoM³eT was the only model applicable to all vision
and vision-language tasks, achieving the best performance on 6 of 12
tasks. In addition, CoM³eT (\(0.442 \pm 0.022\)) outperformed a
theoretical best ensemble using the strongest competing FM for each task
(\(0.357 \pm 0.014\); Fig.~\ref{fig:pretrain-results} A, left).

CoM³eT also achieved the best average performance within each domain. In
pathology (\(0.482 \pm 0.018\); Fig.~\ref{fig:pretrain-results} A,
middle), it was best in 4 of the 6 tasks, including biochemical
recurrence prediction in prostatectomy patients, tumor proportion score
in IHC-stained lung cancer samples, mitotic figure detection in breast
cancer H\&E-stained WSIs, and tissue type segmentation. The runner-up
methods were an ensemble of TITAN\textsuperscript{16} and
CONCH\textsuperscript{17} (\(0.356 \pm 0.015\)), followed by
UNI2\textsuperscript{8} (\(0.348 \pm 0.014\)). In radiology
(\(0.458 \pm 0.027\); Fig.~\ref{fig:pretrain-results} A, right), the
next best method\textsuperscript{18} reached \(0.341 \pm 0.012\) with an
ensemble of two adapted nnU-Net models\textsuperscript{10} pretrained on
CT and MRI. Notably, only CoM³eT solved the multiparametric MRI
segmentation task above random-guessing level, consistent with its
ability to integrate multiple inputs at the same spatial position. An
ensemble of models restricted to 3D patches rather than voxel-level
features performed substantially worse
(\(0.077 \pm 0.006\))\textsuperscript{2,19,20}.

\subsubsection{Multidimensional imaging increases supervised pretraining
data volume}\label{sec:results.scaling}

CoM³eT's first pretraining stage used natural-image datasets to learn
general representations. ImageNet\textsuperscript{21,22} contributed
more than 12 million classification images, and COCO\textsuperscript{23}
added more than 200,000 images for vision-language alignment, object
annotations, and object supercategories. The primary stage then added
medical data from pathology and radiology. Pretraining for pathology
covered more than 10,000 whole-slide images and more than 100,000
annotated patches. In radiology, data ranged from small segmentation
cohorts of a few dozen to a few hundred 3D scans up to large public
resources with thousands of examinations, while meta-learning datasets
such as RadImageNet alone contributed more than \(1.3\) million labeled
images. Taken together, this stage combined supervised data from more
than 100,000 patients across collections spanning single-cell images,
histology patches, whole-slide images, radiographs, CT, endoscopic
images, ultrasound, and MRI.

We quantified the image transformer's performance for new medical
applications via fine-tuning, comparing our multidimensional approach
with averaged 2D processing and, where applicable, a specialized
single-task method.

Pathology tasks quantified the image transformer's effect where each
whole-slide image was represented by many patches. ``Cancer Recurrence
(Whole Slide)''\textsuperscript{24} predicted time to biochemical
recurrence after prostatectomy. CoM³eT achieved an agreement between
predicted and observed event times (C-index) of \(0.736 \pm 0.020\) with
the image transformer and \(0.644 \pm 0.040\) without it. ``Surgery
Outcome (Whole Slide)''\textsuperscript{25} predicted whether
biochemical recurrence would occur within one year after prostatectomy.
Here, performance improved from an AUC of \(0.607 \pm 0.176\) without
the image transformer to \(0.690 \pm 0.152\) with it. Together, these
results show that the image transformer has a strong positive effect on
pathology tasks.

FMs often perform best in data-scarce settings. Here, however, we tested
whether CoM³eT's unified architecture remained competitive with
specialized models on a large task-specific dataset spanning 1,330
imaging studies from more than 10 hospitals\textsuperscript{26}. We
evaluated it on ``Tumor Classification (MRI)'', distinguishing benign
from malignant breast lesions. Among the tested specialized models, a
video Swin Transformer\textsuperscript{27} performed best, with an AUC
of \(0.840\) (95\% CI: \(0.796\) to \(0.879\)). CoM³eT with its image
transformer achieved an AUC of \(0.859\) (95\% CI: \(0.818\) to
\(0.897\)). Without the image transformer, performance dropped to
\(0.782\) (95\% CI: \(0.732\) to \(0.829\)).

\subsubsection{The pyramid transformer extends CoM³eT to
multidimensional segmentation tasks}\label{sec:results.att_segmentator}

When analyzing volumetric images, clinicians rely on context beyond
individual slices. Both nearby slices and global context can indicate
the presence of a structure. For CoM³eT, the pyramid transformer
provided the attention mechanism that combines image patches for
multidimensional segmentation while reusing most parameters with other
multidimensional tasks.

\begin{figure}
\centering
\includegraphics[width=0.8\linewidth,height=\textheight,keepaspectratio,alt={Test data. A, The study included three 3D and two 2D segmentation tasks. Grouped inputs are shown as rows within boxes, and tasks that depend on positional information are connected by a line. B, Four classification tasks were included, one of them ungrouped. C, Synthetic tasks supported model development and ablation studies. D, Segmentation results. Without the image and pyramid transformer (CoM³eT-2D), common problems include discontinuities in tubular structures (red) and missing segments in uncertain regions (purple), other errors remain for CoM³eT (blue).}]{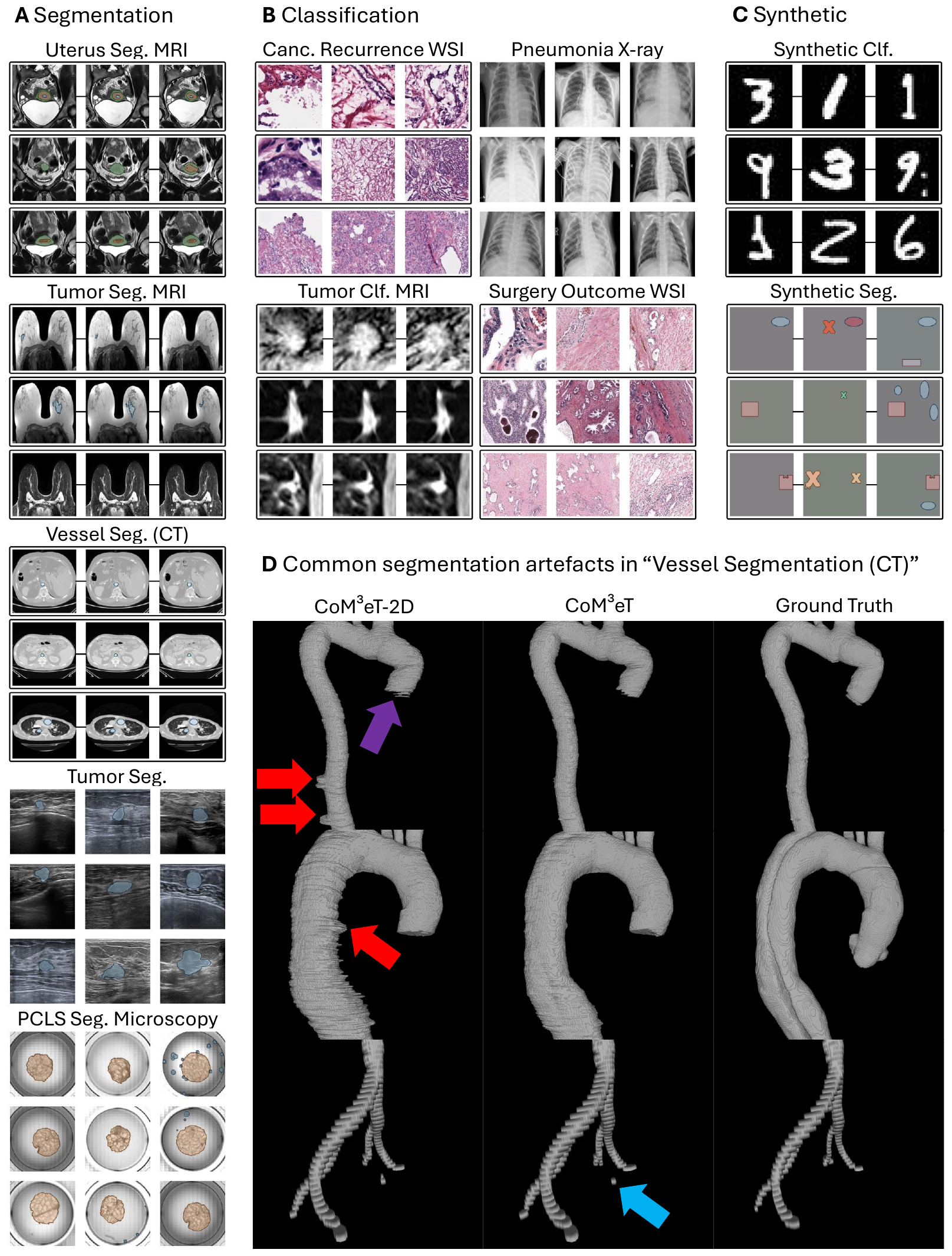}
\caption{\textbf{Test data.} \textbf{A,} The study included three 3D and
two 2D segmentation tasks. Grouped inputs are shown as rows within
boxes, and tasks that depend on positional information are connected by
a line. \textbf{B,} Four classification tasks were included, one of them
ungrouped. \textbf{C,} Synthetic tasks supported model development and
ablation studies. \textbf{D, Segmentation results.} Without the image
and pyramid transformer (CoM³eT-2D), common problems include
discontinuities in tubular structures (red) and missing segments in
uncertain regions (purple), other errors remain for CoM³eT
(blue).}\label{fig:testdata}
\end{figure}

First, synthetic experiments verified that local and global context and
positional information could be used for segmentation
\hyperref[extfig:syndata]{Extended~Data~Fig.~1}. Then, we fine-tuned
CoM³eT on three 3D segmentation settings that differed in dataset size
and context requirements. Representative inputs and task groupings are
shown in Fig.~\ref{fig:testdata}.

On ``Uterus Segmentation (MRI)'', a segmentation task for myometrium,
junctional zone, and endometrium, CoM³eT already exceeded the strongest
specialized baseline without the pyramid transformer
(\(75.49\% \pm 0.5\%\) 3D Dice versus \(71.0\%\) for 3D
U-Net\textsuperscript{28}) on this anisotropic dataset with 13 patients.
Adding the pyramid transformer further improved performance to
\(79.4\% \pm 0.3\%\).

For ``Vessel Segmentation (CT)'', a segmentation task for the aorta and
its branches that was expected to depend strongly on local continuity
across neighboring slices, CoM³eT performed on par with
nnU-Net\textsuperscript{10} even without the pyramid transformer
(\(80.45\%\), 95\% CI: \(79.02\%\) to \(81.47\%\) 3D Dice versus
\(80.38\%\), 95\% CI: \(78.33\%\) to \(81.36\%\)). With the pyramid
transformer, performance increased to \(82.11\%\) (95\% CI: \(80.70\%\)
to \(83.17\%\)), indicating that added multidimensional context improves
even tasks that are challenging for a model that aggregates information
at the patch token level.

The ``Tumor Segmentation (MRI)'' task was to segment breast lesions in
contrast-enhanced MRI using a multicenter dataset of 3,936 patients.
This tested whether CoM³eT's unified architecture remained competitive
outside the data-scarce regime in which FMs often perform best. Overall,
nnU-Net\textsuperscript{10} achieved a \(33.13\%\) 3D Dice score (95\%
CI: \(30.71\%\) to \(35.51\%\)), whereas CoM³eT reached \(25.28\%\)
(95\% CI: \(22.86\%\) to \(27.73\%\)) without the pyramid transformer
and \(58.62\%\) (95\% CI: \(55.92\%\) to \(61.31\%\)) with it. This
large gain suggests that CoM³eT has an advantage when it must both
identify and segment lesions. When we simplified the task to images with
lesions only, Dice scores became similar between methods, at \(53.0\%\)
for nnU-Net (95\% CI: \(50.5\%\) to \(55.4\%\)) and \(53.3\%\) for
CoM³eT (95\% CI: \(50.7\%\) to \(55.9\%\)). Tumor size has traditionally
been summarized by the largest diameter, but automatic segmentation
makes three-dimensional volume practical. The average volume error was
\(4.23\) ml (95\% CI: \(3.57\) to \(4.96\)) for nnU-Net and \(4.46\) ml
(95\% CI: \(3.73\) to \(5.28\)) for CoM³eT without the pyramid
transformer, which the pyramid transformer reduced to \(2.15\) ml (95\%
CI: \(1.65\) to \(2.75\)).

\subsubsection{Partial fine-tuning of CoM³eT is computationally
efficient across various tasks and domains}\label{sec:results.peft}

We compared full and partial fine-tuning on the same tasks and hardware.
For partial fine-tuning, we froze the vision backbone and trained the
other components, including the image transformer, pyramid transformer,
and decoder. This reduced CoM³eT's trainable model size by \(97.5\%\).
It also lowered the time required for an update step at 32 patch tokens
per image from 2330ms to 712ms. Freezing the vision backbone also
allowed caching of patch tokens, which reduced this time by \(>99.9\%\)
compared with full fine-tuning (from \(2.3\) seconds to \(1.5\)ms, see
Fig.~\ref{fig:pretrain-results} C-E).

Across all tasks, an equivalence test supported equivalence between full
and partial fine-tuning (\(p_{\mathrm{lower}} < 0.001\),
\(p_{\mathrm{upper}} = 0.008\)). Neither strategy failed on any of the
tested datasets. Mean performance was similar in both settings, at
\(0.833\) for full fine-tuning and \(0.839\) for partial fine-tuning.

Whole-slide image analysis is a relevant test case for partial
fine-tuning because caching of patch tokens offers large practical
benefits in this setting. For ``Cancer Recurrence (Whole Slide)'',
CoM³eT reached a C-index of \(0.736 \pm 0.020\) with full fine-tuning
and \(0.754 \pm 0.012\) with partial fine-tuning. On ``Surgery Outcome
(Whole Slide)'', CoM³eT reached an AUC of \(0.690 \pm 0.152\) with full
fine-tuning and \(0.795 \pm 0.170\) with partial fine-tuning. Both tasks
showed a trend toward better performance with partial fine-tuning. For
the two whole-slide tasks, we also compared attention topology
augmentation with full attention in the image transformer during partial
fine-tuning. With full attention, CoM³eT reached a C-index of
\(0.757 \pm 0.015\) on ``Cancer Recurrence (Whole Slide)'' and an AUC of
\(0.771 \pm 0.160\) on ``Surgery Outcome (Whole Slide)''. A one-sided
paired t-test did not show that attention topology augmentation
outperformed full attention (\(p = 0.351\)).

Volumetric segmentation was included to quantify partial fine-tuning for
3D segmentation, which relies on both the image transformer and the
pyramid transformer. For the ``Uterus Segmentation (MRI)'' volumetric
segmentation task involving three structures, the myometrium, the
junctional zone, and the endometrium, CoM³eT achieved a mean Dice score
of \(79.4\% \pm 0.3\%\) with full fine-tuning and \(76.3\% \pm 0.4\%\)
with partial fine-tuning, both exceeding the 3D U-Net baseline of
\(71.0\%\)\textsuperscript{28}.

We also evaluated partial fine-tuning on patch-level 2D tasks. Here, the
image transformer and decoder were partially fine-tuned. Because the
group size was one, the image transformer was equivalent to a pretrained
MLP. We evaluated three tasks: ``Pneumonia (X-ray)'', a chest X-ray
classification task, ``PCLS Segmentation (Microscopy)'', a brightfield
microscopy tissue and artifact segmentation task, and ``Tumor
Segmentation (Ultrasound)'', an ultrasound lesion segmentation task. In
``Pneumonia (X-ray)'', which required only weak generalization because
chest X-ray data were included in pretraining, CoM³eT reached an AUC of
\(0.992 \pm 0.002\) with full fine-tuning and \(0.988 \pm 0.003\) with
partial fine-tuning. On ``Tumor Segmentation (Ultrasound)'', which
tested generalization because no ultrasound segmentation data were
included in pretraining, CoM³eT reached a Dice score of
\(75.5\% \pm 2.6\%\) with full fine-tuning and \(71.1\% \pm 2.3\%\) with
partial fine-tuning. The ``PCLS Segmentation (Microscopy)'' task
involved segmentation of tissue and artifacts to enable monitoring of
tissue characteristics over time. Despite including no related data
during pretraining, CoM³eT achieved a mean Dice score of
\(91.8\% \pm 4.7\%\) with full fine-tuning and \(90.0\% \pm 4.5\%\) with
partial fine-tuning.

\subsubsection{Successful federated fine-tuning of CoM³eT in real
multi-site scenarios with consumer-grade hardware and over internet
connections}\label{sec:results.federated-finetuning}

Federated learning followed our partial fine-tuning setup, except that
synchronization occurred every 4 local training steps and each site
updated the model only on its local data subset. This reduced local
compute and synchronization costs (see Fig.~\ref{fig:pretrain-results}
C-E). Keeping the vision backbone frozen allowed caching and reusing
patch tokens during local training, while synchronizing only the
trainable image transformer parameters across sites. We deployed
federated fine-tuning in a three-site study with two German university
hospitals (UKFFM and Charité) and one site contributing public
biochemical recurrence data. The system ran on consumer-grade GPUs over
the internet. Federated fine-tuning started from an earlier version of
CoM³eT which did not include pretraining tasks that used biochemical
recurrence labels. For the centralized baseline (CoM³eT) all data were
pooled.

Model selection was based on the averaged local validation results at
each site (UKFFM: \(0.754\), Charité: \(0.738\), Public: \(0.727\)),
resulting in CoM³eT-FL. CoM³eT-FL achieved a C-index of
\(0.743 \pm 0.012\) on ``Cancer Recurrence (Whole
Slide)''\textsuperscript{24} and an AUC of \(0.820 \pm 0.125\) on
``Surgery Outcome (Whole Slide)''\textsuperscript{25}. For the
corresponding pooled-data CoM³eT achieved a C-index of
\(0.754 \pm 0.012\) and an AUC of \(0.795 \pm 0.170\), respectively. An
equivalence test supported equivalence between CoM³eT-FL and CoM³eT,
ruling out a difference of \(0.05\) or more in either direction
(\(p_{\mathrm{lower}} = 0.019\), \(p_{\mathrm{upper}} = 0.032\)),
suggesting that federated partial fine-tuning can achieve similar
performance to centralized training in these applications.

\subsection{Discussion}\label{discussion}

Traditionally, medical vision has required different model families for
different image and task types. For example, nnU-Net\textsuperscript{10}
has become a general framework for many medical imaging problems, but
its training is restricted to a single segmentation dataset.
Cerberus\textsuperscript{29} supports both classification and
segmentation, but only for 2D histology patches.
CT-FM\textsuperscript{2} incorporates 148,000 scans, but is specific to
3D CT, whereas the million-image-scale model VIRCHOW\textsuperscript{1}
is specific to histology images. CoM³eT, however, has demonstrated that
a single architecture can handle these diverse image and task types,
enabling simultaneous pretraining on standard 2D images and
multidimensional data. This flexibility enabled CoM³eT to scale to
additional pretraining tasks, ultimately winning the UNICORN
competition\textsuperscript{14} for medical FMs, where it was also the
only model that, on its own, was applicable to all vision and
vision-language tasks. However, additional data types and domains
compatible with CoM³eT, such as video-based tasks and optical coherence
tomography, are not yet used.

Previously, foundation models were often pretrained with a single type
of supervision, such as self-supervised learning\textsuperscript{2,8} or
text supervision\textsuperscript{30}. Other approaches use separate
stages to combine multiple levels of supervision. For example,
pathology-specific foundation models\textsuperscript{16,31} commonly
require training a patch encoder as a first stage. However, sequential
training requires the first stage to encode all information useful for
later stages as well. TITAN\textsuperscript{16} then adds more
self-supervised and language-supervised stages, which could also cause
catastrophic forgetting because the same modules are trained
sequentially. For example, fine-tuning the general-purpose model
Gemini\textsuperscript{32} for medical tasks degraded performance on
general tasks, including medically relevant ones such as management plan
appropriateness\textsuperscript{33}. Rather than integrating training
objectives sequentially onto a self-supervised stage, CoM³eT achieves
high-performing representations through a single final pretraining stage
that spans all available data and all levels of supervision, using
structured labels where available and text supervision otherwise. This
challenges the weight the field places on an initial, or even single,
self-supervised pretraining stage. Still, expanding with more weakly and
self-supervised pretraining strategies is a promising future direction.

Segmentation objectives give a clear training signal for CoM³eT's
hyperpixel tokens. These pixel-level representations maintain spatial
context beyond 2D and 3D images by application of the proposed pyramid
transformer, generalizing what task-specific models such as
CAT-Net\textsuperscript{12} achieved with 2D pretrained modules for
anisotropic 3D segmentation. For example, the UNICORN
competition\textsuperscript{14} showed that CoM³eT is the only
multi-purpose model applicable to multiparametric MRI (4D) because it
can process multiple 3D volumes for segmentation. In our own evaluation,
we used vessel segmentation in CT to judge the pyramid transformer's
ability to follow local continuity across neighboring slices.
Qualitatively, segmentation of multiple small vessels on the same slice
remains suboptimal, possibly because the pyramid transformer restricts
attention to slices rather than individual voxels. However, since this
work focused on evaluating CoM³eT's feature quality, we did not optimize
the task heads. For example, using 3D convolutional heads could already
improve performance.

Real-world federated partial fine-tuning of CoM³eT achieved performance
comparable to centralized training, addressing key problems in
developing medical AI: limited access to labeled data, significant
computational requirements, and privacy concerns\textsuperscript{34}.
However, cybersecurity, connection timeouts, and data access policies
posed significant challenges that models cannot address. Without
standardization and trusted training infrastructure, federated learning
will continue to require manual engineering effort and supervision.

A promising next step is to use CoM³eT as the vision component of
conversational assistants. By aligning CoM³eT's embedding space with a
language model, this could address shortcomings in visual performance
that currently limit vision-language models\textsuperscript{33,35,36},
supplying the specialized representations that general models lack.
CoM³eT is equally suited to agentic systems, which rely on external
tools to act on their environment: its auxiliary task heads form a
reusable library of image analysis tools that such systems could combine
into a cohesive care plan\textsuperscript{15}. Because partial and
federated fine-tuning are fast and resource-efficient, this library
could be extended with new vision tools trained on multi-centric data
from clinical information systems, including tasks not covered by the
current pretraining set.

\subsection{Methods}\label{methods}

\subsubsection{Model architecture}\label{sec:methods.architecture}

We used multitask pretraining\textsuperscript{13,37}, in which the model
was structured into shared modules (components used across tasks) and
task heads (components not used across tasks) reflecting increasing
levels of supervision, ranging from language to structured tasks to
pixel-level tasks. Each task consisted of a training objective, a
dataset, a target definition, and a linear task head. A task's target
definition specified what the model predicted (for example a
classification label, a segmentation mask, or a detection target) and
how the loss was computed. The pretraining alternated between all tasks,
forcing the shared modules to learn generally applicable features.

CoM³eT's architecture differed in several ways. For CoM³eT, we adapted
the size of the vision backbone to test its scalability and increased
the number of parameters as shown in Table~\ref{tbl:architecture}. We
also added the image transformer, a transformer module that establishes
multidimensional context from the encoded patch tokens, and the pyramid
transformer module for dense multidimensional segmentation. Together,
these modules allowed CoM³eT to model relationships across image patches
and propagate this context back into dense spatial predictions. Using a
2D vision backbone as a central component also enables to simplify the
architecture for true 2D problems when needed.

Following this design, the vision encoder accepted 2D RGB images with
shape \((3, H, W)\), where \(H\) and \(W\) denote variable spatial
dimensions. The method is compatible with any vision encoder that
supports this input format and produces hierarchical feature maps, such
as
\([(C, \frac{H}{2}, \frac{W}{2}), (2C, \frac{H}{4}, \frac{W}{4}), \ldots]\).
The number of channels \(C\) (e.g., 64, 128, \ldots) depends on the
encoder variant. Here, we selected Swin Transformer
V2\textsuperscript{38} as vision backbone, as these models scale
effectively to large parameter counts and are not restricted to a fixed
image size. To obtain a compact representation for each patch token, we
applied global average pooling followed by a linear projection for
dimensionality reduction. This reduction was especially helpful when
processing medical images composed of many patch tokens. Details on
parameter counts and dimensionality are summarized in
Table~\ref{tbl:architecture}.

\begin{longtable}[]{@{}
  >{\raggedright\arraybackslash}p{(\linewidth - 12\tabcolsep) * \real{0.1405}}
  >{\raggedright\arraybackslash}p{(\linewidth - 12\tabcolsep) * \real{0.0826}}
  >{\raggedright\arraybackslash}p{(\linewidth - 12\tabcolsep) * \real{0.1157}}
  >{\raggedright\arraybackslash}p{(\linewidth - 12\tabcolsep) * \real{0.1074}}
  >{\raggedright\arraybackslash}p{(\linewidth - 12\tabcolsep) * \real{0.1570}}
  >{\raggedright\arraybackslash}p{(\linewidth - 12\tabcolsep) * \real{0.1570}}
  >{\raggedright\arraybackslash}p{(\linewidth - 12\tabcolsep) * \real{0.1983}}@{}}
\caption{\textbf{Architecture statistics.} Each 2D input (such as an
image or image patch) is described by the ``Patch token size'', which
refers to the dimension of the feature vector (patch token) produced by
the encoder for that input. ``Hyperpixel token size'' refers to the
dimension of the hyperpixel token (the embedding corresponding to each
spatial location within the input) as produced by the
decoder.}\label{tbl:architecture}\tabularnewline
\toprule\noalign{}
\multirow{2}{=}{\begin{minipage}[b]{\linewidth}\raggedright
CoM³eT Variant
\end{minipage}} &
\multicolumn{2}{>{\raggedright\arraybackslash}p{(\linewidth - 12\tabcolsep) * \real{0.1983} + 2\tabcolsep}}{%
\begin{minipage}[b]{\linewidth}\raggedright
Vision backbone
\end{minipage}} &
\multicolumn{2}{>{\raggedright\arraybackslash}p{(\linewidth - 12\tabcolsep) * \real{0.2645} + 2\tabcolsep}}{%
\begin{minipage}[b]{\linewidth}\raggedright
\#Parameters transformer
\end{minipage}} &
\multicolumn{2}{>{\raggedright\arraybackslash}p{(\linewidth - 12\tabcolsep) * \real{0.3554} + 2\tabcolsep}@{}}{%
\begin{minipage}[b]{\linewidth}\raggedright
Dimensionality
\end{minipage}} \\
& \begin{minipage}[b]{\linewidth}\raggedright
Variant
\end{minipage} & \begin{minipage}[b]{\linewidth}\raggedright
\#Parameters
\end{minipage} & \begin{minipage}[b]{\linewidth}\raggedright
Image T.
\end{minipage} & \begin{minipage}[b]{\linewidth}\raggedright
Pyramid T.
\end{minipage} & \begin{minipage}[b]{\linewidth}\raggedright
Patch token
\end{minipage} & \begin{minipage}[b]{\linewidth}\raggedright
Hyperpixel token
\end{minipage} \\
\midrule\noalign{}
\endfirsthead
\toprule\noalign{}
\multirow{2}{=}{\begin{minipage}[b]{\linewidth}\raggedright
CoM³eT Variant
\end{minipage}} &
\multicolumn{2}{>{\raggedright\arraybackslash}p{(\linewidth - 12\tabcolsep) * \real{0.1983} + 2\tabcolsep}}{%
\begin{minipage}[b]{\linewidth}\raggedright
Vision backbone
\end{minipage}} &
\multicolumn{2}{>{\raggedright\arraybackslash}p{(\linewidth - 12\tabcolsep) * \real{0.2645} + 2\tabcolsep}}{%
\begin{minipage}[b]{\linewidth}\raggedright
\#Parameters transformer
\end{minipage}} &
\multicolumn{2}{>{\raggedright\arraybackslash}p{(\linewidth - 12\tabcolsep) * \real{0.3554} + 2\tabcolsep}@{}}{%
\begin{minipage}[b]{\linewidth}\raggedright
Dimensionality
\end{minipage}} \\
& \begin{minipage}[b]{\linewidth}\raggedright
Variant
\end{minipage} & \begin{minipage}[b]{\linewidth}\raggedright
\#Parameters
\end{minipage} & \begin{minipage}[b]{\linewidth}\raggedright
Image T.
\end{minipage} & \begin{minipage}[b]{\linewidth}\raggedright
Pyramid T.
\end{minipage} & \begin{minipage}[b]{\linewidth}\raggedright
Patch token
\end{minipage} & \begin{minipage}[b]{\linewidth}\raggedright
Hyperpixel token
\end{minipage} \\
\midrule\noalign{}
\endhead
\bottomrule\noalign{}
\endlastfoot
Tiny & Swin-T & 27,582,570 & 1,054,208 & 529,120 & 256 & 32 \\
UNICORN & Swin-B & 86,905,848 & 4,205,568 & 2,105,472 & 512 & 32 \\
FL & Swin-B & 86,905,848 & 4,205,568 & 2,105,472 & 512 & 64 \\
Base & Swin-B & 86,905,848 & 4,205,568 & 2,105,472 & 512 & 64 \\
Large & Swin-L & 195,220,980 & 16,799,744 & 8,403,904 & 1024 & 128 \\
\end{longtable}

The image transformer was introduced for modeling the interactions and
establishing context between different patch tokens (the fundamental
unit for our training method such as the representation of a 2D image
patch). It generated representations for all patch tokens, which
generalized previous aggregation methods\textsuperscript{11} to
multidimensional medical images for simultaneous processing of dense
prediction tasks such as 3D segmentation and sparse tasks such as
classification. For its architecture, we adopted the general deep
bidirectional encoder of BERT\textsuperscript{39}, which was originally
designed as a pure language model operating over language tokens. Here,
instead of language tokens, we utilized patch tokens generated by the
vision backbone, extending BERT's architecture to process visual
information. Unlike BERT, our pretraining did not require masking input
tokens because each patch token could attend globally. However,
multidimensional medical datasets often contain few cases with many
patch tokens each, which increases the risk of memorizing individual
patients or shortcut features such as pen markings in gigapixel images.
To regularize this setting, we introduced attention topology
augmentation: for multidimensional images, we randomly dropped
connections in the image transformer so that each output patch token
attended only to a random subgroup, reusing the same mask across
attention heads and applying a rate of \(0.5\) for all multidimensional
classification tasks during training and fine-tuning.

Since transformers are permutation-invariant, positional encoding is
required when position is not directly visible in the image. For
volumetric images such as CT and MRI, we chose relative rather than
absolute encoding under the hypothesis that modeling pairwise slice
relationships is sufficient to capture relevant spatial context. For
CoM³eT, we chose ALiBi\textsuperscript{40}, which handles variable
sequence lengths and both short- and long-range dependencies. Because
ALiBi is originally asymmetric, we implemented a symmetric variant so
that the attention bias is the same from position A to B and from B to
A, and extended it to support non-consecutive and duplicate positions
(e.g.~for multiparametric MRI). To remain robust across the different
scales of medical imaging, we encoded distance not in physical units but
as the number of patch tokens between two positions. Thus, adjacent
patch tokens had a distance of 1, while two patch tokens with three
tokens in between had a distance of 4.

\begin{figure}[htbp]
\centering
\includegraphics[width=0.800\linewidth]{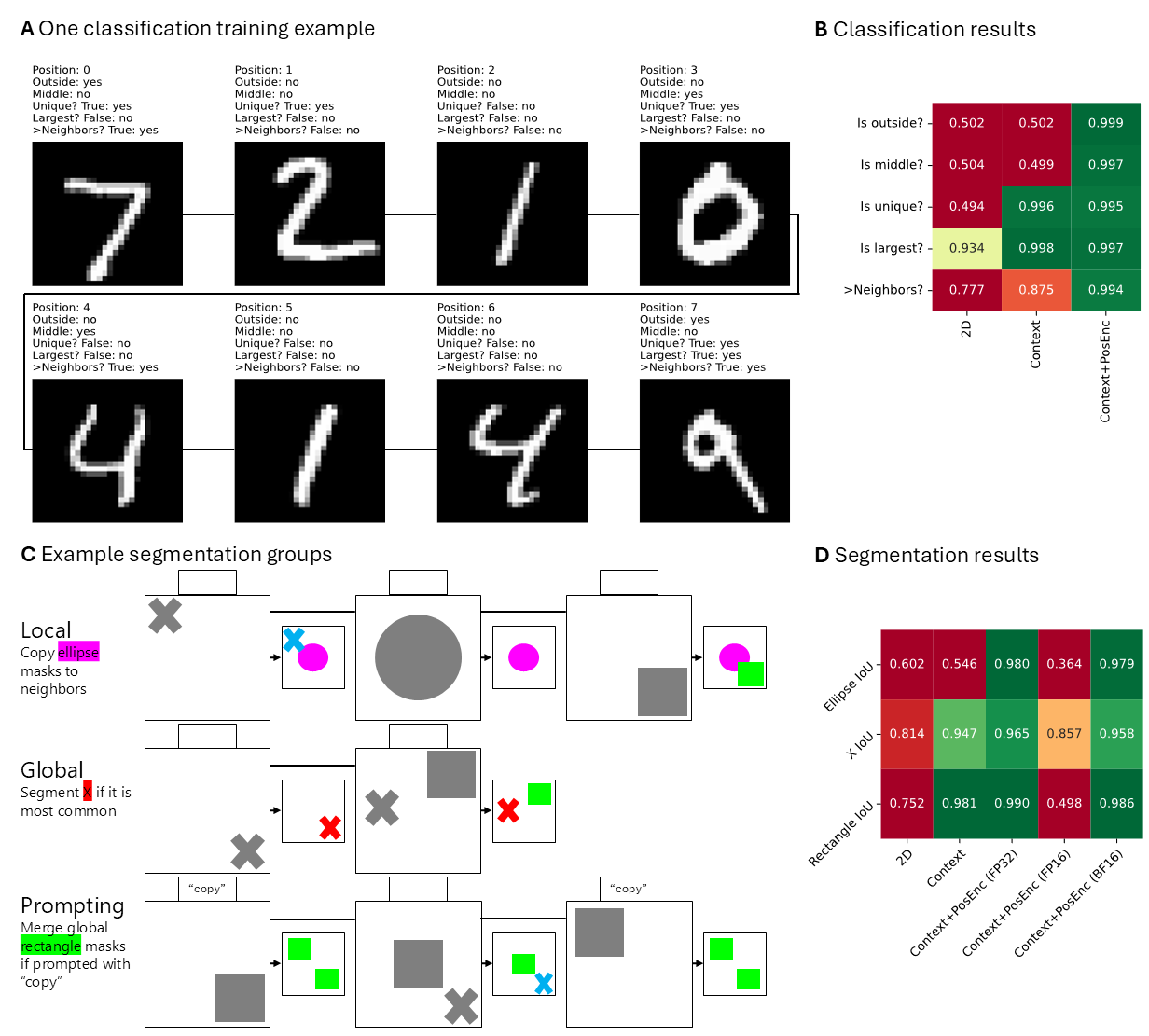}
\caption*{\textbf{Extended Data Fig. 1:} \textbf{Synthetic data. A,} A single example ordered group of 8 images.
\textbf{B,} AUC of a 2D model, a model with the image transformer but
without positional encoding (Context), and the full model with the image
transformer and positional encoding as in CoM³eT (Context+PosEnc).
\textbf{C,} Three example groups. Each image shows the input image
(large square), an optional language prompt (small box), and the
corresponding segmentation mask (small square). The images are shown in
gray so that color indicates the ground-truth segmentation. In this
task, each image contains 0 to 3 objects with varying color, size, and
location, and each group contains 3 to 8 images. \textbf{D,} Performance
of a 2D model (2D), a model with the image transformer and the pyramid
transformer but without positional encoding (Context), and the full
model with additional positional encoding as in CoM³eT (Context+PosEnc).
For the full model, we show results in full precision, half precision,
and bfloat16. Synthetic tasks were used to verify that CoM³eT was able
to learn sparse and dense multidimensional tasks that required both
local and global context, and that ALiBi positional encoding enabled the
model to use position information that is not visible in the images
themselves.}
\phantomsection
\label{extfig:syndata}
\end{figure}

To tune the method and test correctness, we created a synthetic task
from MNIST (28×28 grayscale digit images). For training, we sampled
ordered groups of eight random digits (see
\hyperref[extfig:syndata]{Extended~Data~Fig.~1} A) and added relative
symmetric ALiBi positional encoding between patch tokens. We designed
this task to test both local and global context modeling. For global
context, the model predicted for each image in the ordered group whether
it shows the highest digit in the group and whether any digit is unique
within the group. For local context, it predicted whether an image shows
the highest digit among its two direct neighbors. Because the images
themselves do not indicate position, the model had to rely entirely on
positional encoding. To quantify the expected limitation of relative
encodings for long-range absolute positions such as the middle of the
group, we added two further objectives: whether an image is in the
middle or at the outside of the group. The model was trained for all
learning objectives simultaneously. For synthetic experiments, we used
CoM³eT-Tiny, which used a Swin-Tiny vision backbone with ImageNet-1k
weights\textsuperscript{41}. Its image transformer had 2 layers, 8
attention heads, and a hidden size of 256. All models were trained for
100 epochs. To isolate the effect of context and positional information,
we compared three settings: First, we processed each patch token
independently, with no context. Second, we enabled the image transformer
but disabled positional encoding. This meant that the model could share
information across patch tokens, but could not determine their relative
order or distance. Third, we used both the image transformer and our
symmetric variant of relative ALiBi positional encoding.

We evaluated the image transformer on three representative real-data
sparse tasks, chosen to span different context requirements. In
pathology, ``Cancer Recurrence (Whole Slide)'' and ``Surgery Outcome
(Whole Slide)'' did not use positional encoding and represented each
image by a large number of patch tokens. In radiology, ``Tumor
Classification (MRI)'' used positional encoding. In each case, we
compared CoM³eT against a 2D variant without the image transformer.
Without the image transformer (2D processing), the model operated on
individual patches, so we averaged the result over all patches.

For dense multidimensional prediction, the pyramid transformer used the
output of the image transformer to compute relationships between feature
maps across a multidimensional image. It can be applied wherever a
feature map is computed, however, it is desirable to learn such complex
operations in the shared modules, not in the task-specific modules.

The pyramid transformer took as input a feature map \(F\) of size
\((C, H, W)\) for \(S\) patch tokens and the output \(V\) of the image
transformer of size \(d\) for each patch token, and produced a
contextualized feature map of the same size. First, it computed
relationships \(A\) using only the image transformer's output via
self-attention with sigmoid activation instead of softmax. Here, \(Q\)
and \(K\) were computed from \(V\) via linear layers, with \(Q\) divided
by \(\sqrt{d}\) resulting in the \(S \times S\) relationships
represented by the attention matrix \(A\). Second, \(F'_{i, j}\) for
each spatial location were computed independently based on the feature
maps of all patch tokens:

\[F'_{i, j} = \underbrace{\texttt{sigmoid} \left(QK^T\right)}_{A} F_{i, j}\]

The feature map \(F'\) was further processed by applying a layer
normalization with affine parameters \(\texttt{Norm}\), a GELU
activation, and a learnable layer scale \(\gamma\). Finally, the result
was returned via a residual connection to the input feature map \(F\):

\[\texttt{Output} = F + \gamma \left( \texttt{GELU} \left( \texttt{Norm} \left( F' \right) \right) \right)\]

We used synthetic segmentation tasks to test local context, global
context, and patch token-level prompting. We generated 10,000 training
ordered groups of images and 1,000 test groups, each with 3 to 8 images
of size \(224 \times 224\) and up to 3 objects per image. Each image
showed geometric shapes, including ellipses, rectangles, and X shapes in
different colors. We randomly built the groups once to simulate a
fixed-length dataset and distributed them across multiple images.
Copying masks did not overwrite non-background original masks. The
local-context task required the model to segment ellipses in the image
where they appeared and in the direct neighboring images, where the
ellipse itself was not visible. The global-context task required the
model to segment X-shapes across a group when X-shapes were the most
common object class in that group. The prompting task required the model
to globally copy rectangle segmentations across images that carried a
``copy'' prompt, as illustrated in
\hyperref[extfig:syndata]{Extended~Data~Fig.~1} C. The prompts were
represented by SONAR\textsuperscript{42} text embeddings, and were added
to the patch tokens before the image transformer.

We then evaluated the pyramid transformer on three representative
real-data tasks chosen to span different dataset sizes and context
requirements: ``Uterus Segmentation (MRI)'', ``Vessel Segmentation
(CT)'', and ``Tumor Segmentation (MRI)''. For each task, we compared
CoM³eT against a 2D variant without the transformer components and
against a specialized reference method. The reference was
nnU-Net\textsuperscript{10}, except for ``Uterus Segmentation (MRI)'',
where nnU-Net did not produce useful results and a 3D
U-Net\textsuperscript{28} served as the reference instead. Confidence
intervals were computed by bootstrapping with 10,000 iterations and
sampling with replacement.

\subsubsection{Benchmarking CoM³eT for medical imaging:
UNICORN}\label{sec:methods.unicorn}

In medical imaging, FMs must be applicable to both sparse tasks (for
example, regression and classification) and dense prediction tasks (for
example, detection and segmentation). UNICORN\textsuperscript{14} was a
competition for medical FMs that evaluated frozen models across
pathology and radiology, covering dense and sparse as well as 2D,
multidimensional, vision, and vision-language tasks. Its task set
included 3 classification tasks, 1 regression task, 4 detection tasks, 3
segmentation tasks, and 1 vision-language task, spanning 5 radiology and
7 pathology tasks. Overall, 8 tasks were multidimensional (five 3D and
three WSI), 3 were 2D, and 1 was vision-language. To directly compare
FMs, it used a frozen setting: FM weights remained fixed during
evaluation, and lightweight adapter models were trained on top of
extracted representations. The benchmark used 48 few-shot labeled
training examples for each task. Training and evaluation ran privately
on the Grand Challenge platform.

For our submission, we used CoM³eT-UNICORN. It was identical to
CoM³eT-Base, but used a smaller number of dimensions for the hyperpixel
token because of memory constraints (RAM) on the evaluation platform.
Specifically, we set the dimension of the patch tokens to 512
(unchanged) and the dimension of the hyperpixel tokens to 32. We used
these settings across all tasks and domains.

2D tasks were processed by applying CoM³eT's vision backbone to the
whole image as a single patch token. For each sparse 3D and WSI task
(e.g.~classification), up to 10,000 image patches were extracted as
patch tokens. Because these tasks required a single feature vector per
image, but in our method, the number of feature vectors corresponded to
the number of image patches, we selected the most attended patch token
from the last attention layer of the image transformer as image-level
feature vector and concatenated it with the mean representation of all
patch tokens. For 3D segmentation, it was possible to apply the usual
multidimensional segmentation pipeline of CoM³eT (vision backbone →
image transformer → pyramid transformer → decoder), which produced a 3D
feature map with the same spatial dimensions as the input image and a
feature dimension of 32.

In this paper, we report the performance of CoM³eT on the UNICORN
benchmark, comparing it with the strongest competing model picked for
each task, and with results aggregated by domain (pathology and
radiology) with the second- and third-strongest other models. For more
details on the UNICORN benchmark, please refer to the benchmark
publication\textsuperscript{14}. For confidence intervals (error bars in
Fig.~\ref{fig:pretrain-results} A), we averaged the confidence intervals
of the individual tasks in the same way as the mean task-specific
metrics: rescaled such that the range from random performance to perfect
performance was 0 to 1.

\subsubsection{Pretraining
database}\label{sec:methods.pretraining_database}

Training followed a multistage approach: first natural pretraining based
on large-scale natural image data, then expanded training including both
medical and natural data. Natural image datasets provided a vast amount
of diverse visual training data, which is beneficial for learning
general visual concepts. Medical image datasets contributed general
pretraining data\textsuperscript{4,43} and a diverse collection of
domain-specific knowledge for medical imaging.

To scale the natural-image domain, we used
ImageNet21k\textsuperscript{22} with its more than 11 million labeled
images across roughly 21,000 classes. ImageNet21k is a strong
pretraining dataset but hard to incorporate: its label hierarchy usually
demands specialized handling, or labels to be dropped entirely, as in
Swin V2's self-supervised pretraining\textsuperscript{38}. CoM³eT
instead used all images and labels, addressing label ambiguity by
passing groups of four examples through the image transformer and
predicting the class shared by the group. We also included
ImageNet-1k\textsuperscript{21} for its higher label quality and
established use in pretraining, using its 1,281,167 training images
across 1,000 classes restricted to leaf nodes in the ImageNet hierarchy,
such as ``German shepherd'' rather than broader classes such as
``mammal''. Beyond classification, the database also included natural
images with tasks for vision-language alignment, segmentation, and
object detection from the COCO database\textsuperscript{23},
contributing 118,287 images with 591,753 captions for vision-language
alignment and more than 200,000 images annotated for the 12
supercategories and 80 object categories. For the vision-language task,
negative examples for the discriminator were sampled from other images.

For digital pathology, the database covered patch-level tissue
characterization, tumor detection and grading, mitosis detection, and
dense tissue segmentation across colorectal, breast, and prostate
pathology. CoM³eT was trained with 100,000 H\&E tiles from colorectal
cancer with a classification task over nine tissue classes such as
adipose, lymphocytes and normal mucosa\textsuperscript{44}, 682 gastric
image regions from 155 patients for patch-based segmentation of gastric
mucosa and intestine tissue\textsuperscript{45}, 213 colorectal
adenocarcinoma regions of interest with gland annotations for gland
segmentation\textsuperscript{46}, 4,981 colon regions of interest with
about half a million segmented nuclei\textsuperscript{47}, and 20
whole-slide images with 10 tissue classes for tissue
segmentation\textsuperscript{48}. Regarding breast pathology, the
database included 405 regions of interest with 9,501 mitotic figure
annotations for both detection and segmentation\textsuperscript{49}, 151
whole-slide images with 18 tissue classes such as tumor and
stroma\textsuperscript{50}, and 370 whole-slide images with seven
classes such as invasive tumor, in-situ tumor, and inflamed
stroma\textsuperscript{51}. For ovarian pathology, tumor subtype patch-
and image-level annotations were included\textsuperscript{52}. In
prostate pathology, it included more than 10,000 whole-slide images for
tumor grading and tissue-type segmentation\textsuperscript{53}, around
2,000 patients for biochemical recurrence prediction from
public\textsuperscript{54,55} and private sources, epithelium
segmentation\textsuperscript{56}, basal-cell segmentation (private),
prostate histopathology classification with Gleason
grading\textsuperscript{57,58}, and about 200,000 additional tiles for
tumor detection\textsuperscript{59,60}. Other data included captions
mined from pathology publications, included as a vision-language
task\textsuperscript{4}, and tumor proportion score prediction for IHC
stained lung tissue\textsuperscript{61}.

Because CoM³eT processes stain types beyond H\&E, foreground extraction
through simple thresholding was not possible. However, sampling within
tissue foreground tiles during pretraining was necessary for efficiency.
Furthermore, we wanted to exclude pen annotations and regions with heavy
artifacts. We used the partial fine-tuning setting with earlier versions
of CoM³eT, and iteratively fine-tuned a multitask model for foreground
and artifact detection. In each iteration, we annotated one image,
preferably by correcting a model mistake while leaving most of the image
unlabeled, then partially fine-tuned the model for a few seconds using
25 update steps with all annotated data collected so far, and finally
predicted on another random whole-slide image to identify where the next
annotation was needed. The final foreground model was trained with 147
sparsely annotated zoomed out images and used for sampling of foreground
regions in all histology datasets.

When loading a tomographic image, training used 3D patches with a
probability of 80\%, and otherwise full volumes. Patches had a randomly
chosen depth between 1 and 32 slices and were sampled such that the
corresponding volume contained at least one foreground voxel when
possible. The patch edge length depended on the task and was most
commonly between 128 and 384 pixels. To keep data loading from becoming
a bottleneck in the main training process, we used group caching with an
adaptive cache rate. We first loaded a group of image patches, applied
augmentations that required the whole group, including random 3D
rotations, resampling, cropping, scaling, and intensity augmentations,
and then cached the result. We then loaded a subgroup from the cache and
applied local augmentations, including rotations in 90 degree steps,
additional intensity augmentations, and distortions. To pretrain the
model to extract general features in radiology,
RadImageNet\textsuperscript{43} was used. This dataset contributes over
one million images of various anatomical structures and pathologies.

For multidimensional segmentation, the pretraining database included
TotalSegmentator\textsuperscript{62}, a collection of 3D CT segmentation
masks for 24 organs, 26 vertebrae, 18 cardiac structures, 23 muscles,
and 26 ribs with five segmentation tasks (one per category), and the
Medical Segmentation Decathlon\textsuperscript{63} with 10 3D
segmentation tasks. Further, it included anatomic structure
segmentations for prostate\textsuperscript{64}, abdominal organs in
CT\textsuperscript{65}, spinal structures in MRI\textsuperscript{66}.

For lesion-related pretraining, we used public data with 32,120 lesions
using rectangle annotations with an object detection task in
CT\textsuperscript{67}, 7,095 regions of interest around lesions with a
segmentation task for CT\textsuperscript{68}, segmentation masks of
1,507 patients for breast MRI\textsuperscript{69} and 369 patients for
brain tumor in MRI\textsuperscript{70}. We also included private
full-body CT lesion segmentation from a melanoma
cohort\textsuperscript{71,72}, comprising 4,727 lesions from 262
patients with stage IV malignant melanoma, including lung (2,006), liver
(753), soft tissue or skin (1,155), lymph node (379), skeletal (123),
spleen (99), and other lesions (212). Lesion malignancy classification
data were included for breast MRI\textsuperscript{73}, and
CT\textsuperscript{74,75}.

Further patch-level data broadened the database. For chest radiography,
we included anatomical segmentation\textsuperscript{76}, pneumothorax
segmentation\textsuperscript{77}, and identification of common thoracic
findings\textsuperscript{78,79}. Beyond radiology, the database covered
cell-level classification of bone marrow cytology in hematologic
malignancies\textsuperscript{80} and polyp segmentation in
gastrointestinal endoscopy\textsuperscript{81}.

\subsubsection{Training method}\label{sec:methods.training_method}

The training method was built on UMedPT\textsuperscript{13} with key
changes. Because pretraining alternates between many tasks, keeping
every task head and its gradients in memory would make the memory
footprint grow with the number of tasks. Instead, each step processed
one task at a time: after a task's turn, we applied the update to its
head and immediately discarded the task-specific gradients before moving
to the next task. As a result, memory requirements stayed constant
enabling a larger batch size and more tasks with the same hardware.

Tasks with structured labels such as classification used a single linear
layer, and segmentation tasks used a single convolutional layer. Besides
segmentation and classification, CoM³eT supported vision-language
pretraining. For this purpose, we used SONAR\textsuperscript{42} as text
model, which can both compute sentence embeddings and generate text from
embeddings. Captions were augmented using the Qwen-3 chat
model\textsuperscript{82}. The training objective was to predict the
caption from the image and discriminate correct captions from incorrect
ones, inspired by CoCa\textsuperscript{83}. Like CoCa, our approach was
trained for both captioning and caption discrimination. Unlike CoCa,
which jointly trains the text decoder and vision encoder, we aligned
with a frozen pretrained unimodal text decoder to free GPU memory and
avoid catastrophic forgetting. Our primary objective was pretraining and
not captioning performance, so we used a high loss weight for caption
discrimination (80\%) and only a low weight for caption generation
(20\%).

For CoM³eT's vision backbone, we found that fp16 produced NaN losses
with the Swin Transformer, which led to the poor performance shown in
\hyperref[extfig:syndata]{Extended~Data~Fig.~1}, and therefore used
bfloat16. Together, mixed precision and gradient checkpointing enabled
training with 2× larger batch sizes, which was important for processing
large multidimensional images with CoM³eT-Large. CoM³eT-Large and Base
variants used bfloat16. CoM³eT-Tiny used full precision (fp32).

For multi-node training, we used the ZeRO optimizer\textsuperscript{84},
which partitions optimizer state across processes so that no single
device holds a full copy. This reduces per-device memory relative to
standard Adam, whose optimizer states require roughly twice the model
size. Each node processed a subset of tasks, accumulated local
gradients, and averaged them across nodes via PyTorch's
DistributedDataParallel using NCCL for inter-GPU communication.
Auxiliary task heads were kept fully local, while shared model
components were managed by DDP. Training ran on 10 A100 80GB GPUs across
five machines, each with 334 GB RAM and 24×2.20 GHz Intel Xeon Gold 5320
CPUs.

To train with many large datasets, several technical advancements were
implemented. Gradient checkpointing was used around each block of the
vision backbone. Two variants of gradient checkpointing were tested to
reduce memory requirements during training: one checkpoint around the
entire vision backbone, which has the advantage of being simple to add
to any architecture, and one checkpoint around each encoder segment as
done in\textsuperscript{38}, which is more memory efficient but requires
more implementation effort when using other types of encoders. Both
variants were tested with synthetic data for classification,
segmentation, group classification, and group segmentation tasks.

\subsubsection{Partial fine-tuning of CoM³eT}\label{sec:methods.peft}

We explored partial fine-tuning as a parameter-efficient fine-tuning
strategy for CoM³eT. Specifically, we froze the vision backbone and
trained the image transformer, pyramid transformer, and decoder modules.
For patch-level tasks, the image transformer was equivalent to a
pretrained MLP because the group size was one, and classification tasks
did not use the decoder. The resulting patch tokens were cached once and
reused during training.

We compared full and partial fine-tuning. To do so, we selected
representative downstream tasks that covered the main task types.
``Uterus Segmentation (MRI)'' was chosen for 3D segmentation, and
``Surgery Outcome (Whole Slide)'' and ``Cancer Recurrence (Whole
Slide)'' were chosen for gigapixel image analysis. For 2D tasks,
``Pneumonia (X-ray)'' represented classification and ``Tumor
Segmentation (Ultrasound)'' and ``PCLS Segmentation (Microscopy)''
represented segmentation. ``Uterus Segmentation (MRI)'', ``Tumor
Segmentation (Ultrasound)'' and ``PCLS Segmentation (Microscopy)''
tested for strong generalization because no similar data were included
in pretraining.

Each update step sampled eight patients for multidimensional tasks and
64 patients for patch-based tasks. All trainings used 15 loops with 100
update steps each. Up to 2,500 patch tokens per patient were cached.
Because full fine-tuning is memory-intensive and we wanted identical
settings across both strategies, we subsampled 50 to 100 tokens per
patient per update step whenever more were available. Unless stated
otherwise, we used cross-validation with five splits of 70\% training
and 30\% validation data, repeated three times, yielding 15 samples per
experimental condition.

\subsubsection{Federated fine-tuning of
CoM³eT}\label{sec:methods.federated-finetuning}

We implemented federated learning with Flower\textsuperscript{85} to use
CoM³eT with federated fine-tuning for a single downstream application.
The FM in this experiment was CoM³eT-Base without biochemical
recurrence-specific pretraining data. Starting from this pretrained
checkpoint, we performed federated fine-tuning to obtain CoM³eT-FL. As
in the partial fine-tuning experiment, we fine-tuned the image
transformer while keeping the vision backbone frozen, and each site kept
its patient data local and shared only updates for the trainable
parameters. We used a standard server-based optimization scheme similar
to Federated Averaging\textsuperscript{6}: Each site trained locally for
a short interval, sent parameter updates for aggregation, and received
the updated global model from the server. Because the transferable part
of the model was small, we synchronized after every 4 local update steps
rather than after full local epochs.

We ran this setup in a real multi-site scenario with three sites: two
German university hospitals and a research institute that pooled
relevant public data. The sites used consumer-grade GPUs and weak
internet connections under real-world hospital IT constraints.
Communication between nodes used gRPC over HTTP/2. We used a
containerized AI environment that isolated both patch token caching and
federated fine-tuning from the host system and allowed each participant
to define separate read and write storage resources to meet
data-protection requirements. For patch-token caching, Charité used four
Tesla T4 GPUs, and UKFFM used four RTX A5000 GPUs. Federated fine-tuning
then ran on a single Tesla T4 GPU at Charité and a single RTX A5000 GPU
at UKFFM.

\subsubsection{Statistics}\label{sec:methods.statistics}

To compare full and partial fine-tuning, we used equivalence testing
with two one-sided tests (TOST) between task and training-split pairs,
ruling out that one strategy was meaningfully worse than the other by a
predefined margin of \(0.05\) for the main performance metric of each
task (for example, AUC for classification and C-index for time-to-event
prediction). Each side was a Wilcoxon signed-rank test on the paired
differences. We considered two strategies equivalent if both
\(p_{\mathrm{lower}}\) and \(p_{\mathrm{upper}}\) were below \(0.05\),
and used the same test to compare federated with centralized
fine-tuning. To assess whether attention topology augmentation
outperformed full attention when fine-tuning, we used a one-sided paired
t-test, pairing observations by task and cross-validation split across
the two whole-slide tasks (\(n=10\) pairs), and verified the normality
assumption of this test with a Shapiro-Wilk test on the paired
differences.

\subsubsection{Test data}\label{test-data}

\paragraph{Cancer Recurrence (Whole
Slide)}\label{sec:methods.testdata.cancer-recurrence-whole-slide}

``Cancer Recurrence (Whole Slide)'' is a time-to-event prediction task
for biochemical recurrence after prostatectomy. The quality of learned
representations was evaluated with PRAD multi-center
data\textsuperscript{24} split into five centers. Four centers (N=97,
N=86, N=85, N=71) were kept as is, while the remaining centers, which
often contained only one sample each, were pooled into one center (220
samples from 27 sites). This task was included to enable the comparison
of models for their application on datasets with many patch tokens per
patient.

\paragraph{Surgery Outcome (Whole
Slide)}\label{sec:methods.testdata.surgery-outcome-whole-slide}

``Surgery Outcome (Whole Slide)'' is a whole-slide image classification
task that predicts whether biochemical recurrence occurs within one year
after prostatectomy based on the CHIMERA challenge
data\textsuperscript{25}. It contains data from 94 patients with 190
images. This task complemented ``Cancer Recurrence (Whole Slide)'' to
compare models for whole-slide image analysis with many patch tokens per
patient.

\paragraph{Uterus Segmentation
(MRI)}\label{sec:methods.testdata.uterus-segmentation-mri}

``Uterus Segmentation (MRI)'' is a volumetric segmentation task for
three uterine structures: myometrium, junctional zone, and endometrium.
All structures were segmented up to the cervical junction by a
radiologist. The dataset is part of the RACOON FADEN project on early
detection of adenomyosis and comprises 13 female subjects
(\(27 \pm 5.44\) years, BMI \(21.8 \pm 2.96\)) collected across three
German university hospitals. T2-weighted short-axis uterine MRI were
acquired with motion-insensitive, multi-shot TSE BLADE sequences,
yielding an anisotropic voxel spacing of
\(0.643 \times 0.643 \times 3.0\) mm. Methods were compared over five
repetitions on the same test patients used in the dataset's
publication\textsuperscript{28}, and compared with a 3D U-Net as the
baseline method. nnU-Net was also evaluated but did not produce useful
results. This task was included to compare models for 3D segmentation in
a small-data setting, where FMs are expected to be particularly useful.

\paragraph{PCLS Segmentation
(Microscopy)}\label{sec:methods.testdata.pcls-segmentation-microscopy}

``PCLS Segmentation (Microscopy)'' assessed the segmentation of
precision-cut lung slice (PCLS)\textsuperscript{86} tissue samples in
brightfield microscopy images. PCLS are an ex vivo organotypic model
prepared by slicing agarose-inflated lung tissue with a vibratome,
preserving the three-dimensional cellular architecture, resident cell
populations, and cell--matrix relationships of the native lung. Recent
advances in culture conditions have enabled long-term (multi-week)
cultivation of viable PCLS, opening the door to longitudinal monitoring
of tissue behavior over extended time periods. This, in turn, permits
the analysis of fibrotic tissue remodeling and treatment response over
time\textsuperscript{87,88}. For such applications, accurate
segmentation of the tissue piece itself is the first processing step, as
it enables analyses of relative tissue growth or shrinkage over time (a
potential marker of fibrosis progression) as well as masking of
subsequent tissue analyses. Tissue segmentation in these microscopy
images is non-trivial due to various challenges such as recording
artifacts (e.g.~poorly focused recordings), ambiguous tissue boundaries,
shadows, and formation of air bubbles. The PCLS microscopy recordings
were obtained using a Zeiss Axio Observer Z1 brightfield microscope by a
trained laboratory technician. For tissue segmentation, recordings were
downscaled to 512×512 and converted to JPGs. Tissue and background
partial annotations were created using LabelStudio for 86 sample
recordings. This task was used to assess the ability of partial
fine-tuning to generalize to a new modality.

\paragraph{Tumor Segmentation
(Ultrasound)}\label{sec:methods.testdata.tumor-segmentation-ultrasound}

``Tumor Segmentation (Ultrasound)'' is a 2D lesion segmentation task
consisting of 256 B-mode ultrasound scans from 256
patients\textsuperscript{89}. Of these scans, 154 show benign tumors, 98
show malignant tumors, and four show normal tissue. Each scan was
manually segmented by five experienced radiologists. The dataset is
skewed towards smaller tumors, supporting the development of methods for
earlier detection. This task tested generalization because no
ultrasound-specific pretraining data was included.

\paragraph{Pneumonia (X-ray)}\label{sec:methods.testdata.pneumonia-xray}

``Pneumonia (X-ray)''\textsuperscript{90} is a chest X-ray
classification task for pneumonia diagnosis containing 5856 pediatric
images, each labeled as either normal or pneumonia. It was included as
the 2D classification task in the partial fine-tuning experiments and
enabled this study to test weak generalization, because while no
pediatric images were included, chest X-ray data were already included
in pretraining of CoM³eT.

\paragraph{Vessel Segmentation
(CT)}\label{sec:methods.testdata.vessel-segmentation-ct}

``Vessel Segmentation (CT)'' is a binary segmentation task for the aorta
and its branches in 3D CTA images, selected as a 3D setting expected to
depend strongly on local continuity across neighboring slices. Training
data\textsuperscript{91} consisted of 100 CTA images with axial
dimensions ranging from 389×389 to 516×516 pixels (average 450×450),
isotropic voxel resolution of 1 mm, and 578 to 801 axial slices per
image (average 695). Test data\textsuperscript{92} comprised 56 CTA
scans from mostly healthy aortas, covering the aortic arch and its
branches and the abdominal aortas with the iliac arteries, with expert
binary segmentation masks. The collection included one case with an
abdominal aortic aneurysm and five cases with aortic dissections.
nnU-Net\textsuperscript{10} was used as the baseline method. This task
was included to compare models for a task that requires volumetric
continuity.

\paragraph{Tumor Segmentation
(MRI)}\label{sec:methods.testdata.tumor-segmentation-mri}

``Tumor Segmentation (MRI)'' is a 3D segmentation task for breast
lesions in dynamic contrast-enhanced MRI, covering malignant, benign,
and lesions of unknown histology with a mean lesion volume of
\(6.1 \pm 20.3\) ml. The MRI protocols included acquisitions with and
without fat suppression on 1.5T and 3T scanners from GE, Siemens
Healthineers, and Philips Healthcare, with varying voxel sizes and
fields of view. nnU-Net\textsuperscript{10} was used as the baseline
method. This task was included to compare the 3D segmentation
capabilities of models for which multi-center robustness and data
efficiency are irrelevant due to the inclusion of a large dataset from
multiple hospitals.

\paragraph{Tumor Classification
(MRI)}\label{sec:methods.testdata.tumor-classification-mri}

The ``Tumor Classification (MRI)'' task distinguished between benign and
malignant breast lesions in dynamic contrast-enhanced MRIs based on data
from a private Siemens Healthineers collection. The collection spanned
more than ten clinical institutions, including 1,330 imaging studies
from 1,299 patients and 1,796 lesions with an average volume of 4.8 ±
14.8 ml. The malignant class included lesions that were histologically
confirmed as malignant or judged as obviously malignant by a radiologist
(BI-RADS 5). The benign class included lesions that were histologically
confirmed as benign or obviously benign (BI-RADS 2) and therefore not
biopsied. A video Swin Transformer\textsuperscript{27} trained on small
regions of interest around the lesions was selected as the baseline
method due to its high performance, multidimensional pretraining based
on the Kinetics video dataset, and open-weight
availability\textsuperscript{41}. This large, multicenter dataset was
used to create a task that favors specialized, single-task baselines for
3D classification, testing whether CoM³eT's unified architecture would
remain competitive.

\subsection{Data availability}\label{data-availability}

The following test datasets are publicly available: ``Cancer Recurrence
(Whole Slide)''\textsuperscript{24}, ``Surgery Outcome (Whole
Slide)''\textsuperscript{25}, ``Vessel Segmentation
(CT)''\textsuperscript{91,92}, ``Tumor Segmentation
(Ultrasound)''\textsuperscript{89}, and ``Pneumonia
(X-ray)''\textsuperscript{90}. The following test datasets can be
obtained from the corresponding author upon reasonable request: ``Uterus
Segmentation (MRI)'' with additional permission from the RACOON
consortium, ``Tumor Segmentation (MRI)'' and ``Tumor Classification
(MRI)'' with additional permission from Siemens Healthineers, and ``PCLS
Segmentation (Microscopy)'' with additional permission from Fraunhofer
ITEM.

\subsection{Code availability}\label{code-availability}

The code is publicly available at
\url{https://github.com/FraunhoferMEVIS/MedicalMultitaskModeling}.

\subsection{References}\label{references}

\protect\phantomsection\label{refs}

\subsection{Acknowledgements}\label{acknowledgements}

This study was funded by the German Federal Ministry of Research,
Technology, and Space (BMFTR; PROSurvival, funding code 01KD2213A-D \&
NUM 2.0, funding code 01KX2121), the Deutsche Forschungsgemeinschaft
(DFG, German Research Foundation) (SPP2177 program, project number
428216905 and SFB 1382 The Gut-Liver Axis, project number 403224013, and
RTG2375, project number 331065168).

\subsection{Author contributions}\label{author-contributions}

J.R.S. prepared the test data, together with K.G., H.v.B. and R.G.
(Tumor Segmentation (MRI) and Tumor Classification (MRI)); C.T., M.M.,
L.S. and S.A. (Uterus Segmentation (MRI)); K.H. (Vessel Segmentation
(CT)); and E.P., A.P. and N.A. (PCLS Segmentation (Microscopy)). J.R.S.,
K.G., K.H. and E.P. analysed the results. J.R.S., T.N., T.B., I.Do.,
R.M., N.F., P.W. and N.Z. ran the federated-learning experiments. J.R.S.
and H.M. built the pretraining infrastructure, and J.R.S. and N.W.
developed the software for data curation. J.R.S., T.N., L.O.S., A.G.,
J.H.M., F.P., I.Da., T.R.K. and S.E. prepared the pretraining data.
P.W., T.R.K. and S.E. contributed as clinical advisors for pathology,
and F.K., F.P. and I.Da. as clinical advisors for radiology. J.R.S.
wrote the paper, with feedback from all authors. F.K. and J.L.
contributed equally and coordinated the study.

\subsection{Competing interests}\label{competing-interests}

The applicant `Fraunhofer-Gesellschaft zur Förderung der angewandten
Forschung eingetragener Verein' has a patent pending related to the
training algorithm and neural architecture components presented in this
article (patent application no. EP23209015.9; names of inventors,
J.R.S., T.N., J.L., F.K.).

\pagebreak

\subsection{Supplementary information}\label{supplementary-information}

\subsubsection{Partial fine-tuning of CoM³eT performs competitively in
predicting biochemical recurrence in prostate
cancer}\label{sec:supplementary.leopard}

We submitted an early version of CoM³eT to the LEOPARD
study\textsuperscript{54} using partial fine-tuning. This model had been
pretrained on 42 tasks including prostate-cancer-related tasks such as
ISUP grading and tumor detection in whole-slide images but did not
include biochemical-recurrence-specific data. We then partially
fine-tuned it in a multitask setting that included the LEOPARD training
data and the PRAD cohort\textsuperscript{24}, each with its own task
head. The study evaluated four cohorts of varying size, each contributed
by a single hospital, and compared other FMs and specialized algorithms
in this external setting.

CoM³eT achieved the best result on a single private cohort with a
C-index of \(0.771\) (95\% CI: \(0.727\) to \(0.813\)). The second-best
result was \(0.714\) (95\% CI: \(0.669\) to \(0.761\)) based on the
pathology FM CTransPath\textsuperscript{93}. Across all four cohorts,
with \(N=100\) test patients from the training cohort and \(N=723\),
\(N=427\), and \(N=332\) from the external cohorts, CoM³eT reached a
C-index of \(0.707\), tied with tumor grading as currently used in
clinical practice. When weighted by the number of test patients in each
cohort, CoM³eT achieved a C-index of \(0.720\), ahead of the next best
results of other participants with \(0.704\) and \(0.699\).

\end{document}